# A locally statistical active contour model for SAR image segmentation can be solved by denoising algorithms

Guangming Liu
Yantai Science and Technology Association, email:1733823365@qq.com

**Abstract**─In this paper, we propose a novel locally statistical variational active contour model based on I-divergence-TV denoising model, which hybrides geodesic active contour (GAC) model with active contours without edges (ACWE) model, and can be used to segment images corrupted by multiplicative gamma noise. By adding a diffusion term into the level set evolution (LSE) equation of the proposed model, we construct a reaction-diffusion (RD) equation, which can gradually regularize the level set function (LSF) to be piecewise constant in each segment domain and gain the stable solution. We further transform the proposed model into a general ROF model by adding a proximity term ,and it can be solved by a fast denoising algorithm proposed by Jia-Zhao or soved by BM3D and NLM denoising algorithm, which also provide a unified solution framework for formally generalized-ROF-like subproblems arising in multivariate splitting algorithms. [27] was submitted on 29-Aug-2013, and our early edition was ever submitted to TGRS on 12-Jun-2012, Venkatakrishnan et al. [31] proposed their **"PnP algorithm"** on 29-May-2013, so Venkatakrishnan and we proposed the **"PnP algorithm"** almost simultaneously. In this paper, we propose two fast fixed point algorithms to solve SAR image segmentation question. Experimental results for real SAR images show that the proposed image segmentation model can efficiently stop the contours at weak or blurred edges, and can automatically detect the exterior and interior boundaries of images with multiplicative gamma noise. The proposed FPRD1/FPRD2 models are about 1/2 (or less than) of the time required for the SBRD model based on the Split Bregman technique.

## Ⅰ.INTRODUCTION

Segmentation technique is a crucial step in synthetic aperture radar images automatic interpretation. Due to the presence of speckle, segmentation of SAR images is generally acknowledged as a difficult problem. Recently, level set methods (LSM) have been extensively applied to SAR images segmentation [1]-[4]. We know that level set methods can be divided into two major categories: region-based models [1-8] and edge-based models [9-12]. In order to make use of edge information and region characteristic, region-based model is often integrated with edge-based model to form the energy functional of models. There are some advantages of level set methods over classical image segmentation methods, such as edge detection, thresholding, and region grow: level set methods can give sub-pixel accuracy of object boundaries, and can be easily formulated under a energy functional minimization, and can provide smooth and closed contour as a result of segmentation and so forth.

However, in conventional level set formulaions, the level set function typically develops irregularities during its evolution, which may cause numerical errors and eventually destroy the stability of the evolution. Therefore, a numerical remedy, called reinitialization, is typically applied to periodically replace the degraded LSF with a signed distance function. However, the practice of reinitialization not only raises serious problems as when and how it should be performed, but also affects numerical accuracy in an undesirable way.

In recent years, some variational level set formulations[7], have been proposed to regularize the LSF during evolution, and hence the re-initialization procedure can be eliminated. These variational LSMs without re-initialization have many advantages over the traditional methods.

Kaihua Zhang propose a new LSM, namely the RD method[13], which is completely free of the costly re-initialization procedure. The RD method has the following advantages over the traditional level set method and state-of-the-art algorithms. First, the RD method has much better performance on weak boundary anti-leakage. Second , the implementation of the RD equation is very simple and it does not need the upwind scheme at all. Third, the RD method is robust to noise.

However, a common difficulty with above variational image segmentation models [1-12] is that the energy functionals to be minimized are all not convex. Thus we may obtain a local minimizer rather than a unique global minimizer from these non-convex energy functionals. This is a more serious problem because local minima of image segmentation often has completely wrong levels of detail and scale. In order to resolve the problems associated with non-convex models, Chan-Esedoglu-Nikolova [14] proposed a convex relaxation approach for image segmentation models. But the globally convex segmentation models contain a total variation (TV) term that is not differentiable in zero, making them difficult to compute. It is well known that the split Bregman method is a technique for fast minimization of $l^1$ regularized energy functional. Goldstein-Osher (GO) [15-15] proposed a more efficient way to compute the TV term by applying the Split Bregman method. Recently, Jia-Zhao[17-18] gave a fast algorithm for the solution of ROF denoising model [19] based on anisotropic TV term, which is much faster than GO algorithm.

On the one hand, we know SAR images are affected by speckle, a multiplicative gamma noise that gives the images a grainy appearance and makes the interpretation of SAR images a more difficult task to achieve. Recently, Steidl and Teuber [20] examined theoretically and numerically the suitability of the I-divergence-TV convex denoising model for restoring images contaminated by multiplicative Gamma noise, which is the typical data fitting term when dealing with Poisson noise. Inspired by [5], we propose a novel variational active contour model based on I-divergence-TV model for SAR image segmentation. On the other hand, it is well known that the ROF denoising model is strictly convex, which always admits a unique solution. By establishing the equivalence relationship between the energy functional of image segmenatation and the weighted ROF denoising model, we further give two fast fixed point algorithms based the algorithm proposed by Jia-Zhao which do not involve partial differential equations.

This paper is organized as follows. Section Ⅱ describes the proposed locally statistical variational active contour model. Section Ⅲ describes the proposed two fast globally convex segmentation algorithms, and Section Ⅳ describes the experimental results. Section Ⅴ concludes this paper.

## Ⅱ. A NOVEL LOCALLY STATISTICAL VARIATIONAL ACTIVE CONTOUR MODEL BASED ON AA MODEL

### A. a novel locally statistical variational active contour model based on I-divergence-TV .

We denote by $R^2$ the usual 2-dimensional Euclidean space $H$. We use $\langle .,. \rangle$ and $\| \|$, respectively, to denote the inner product and the corresponding $l^2$ norm of an Euclidean space $H$ while $\| \|_1$ is used to denote the $l^1$ norm. We denote by $\nabla_x^T$ ($\nabla_y^T$) the conjugate of the gradient operator $\nabla_x$ ($\nabla_y$).

Given a observed intensity image $f : \Omega \to R (f > 0)$, where $\Omega$ is a bounded open subset of $R^2$, speckle is well modeled as a multiplicative random noise $n$, which is independent of the true image $u$, i.e. $f = u \cdot n$. We know that fully developed multiplicative speckle noise is Gamma distributed with mean value $\mu_n = 1$ and variance $\sigma_n{}^2 = 1/L$, where $L$ is the equivalent number of independent looks of the image. G.Steidl and T.Teuber [20] proposed a convex energy functional to remove multiplicative gamma noise, i.e.

$$E(u) = \int_\Omega |\nabla u| dx + \mu \int_\Omega (u - f \log u) dx . \tag{1}$$

The first term of (1) is the TV regularization term, and the second term is the data fidelity term, and $\mu$ is a positive constant parameter to balance the first term and second term. The distinctive feature of the TV regularization term and its various variants is that edges of images are preserved in the denoised image. We adopt the idea of two-phase segmentation question model [5][7], and suppose that the true image $u$ is piecewise constant, i.e. $u = C_1$ for $x \in \Omega_1$, and $u = C_2$ for $x \in \Omega_2$, where $\{\Omega_1, \Omega_2\}$ is a partitioning of the image domain $\Omega$. Using the level set formulation, the true image $u$ can be expressed：

$$u = C_1 H(\phi) + C_2(1 - H(\phi)). \tag{2}$$

Where $\phi$ is a level set function, $H(\phi)$ is the Heaviside function. We then propose a novel locally statistical variational active contour model based on I-divergence-TV [28], the energy functional can be written in the level set formulation as:

$$E(\phi, C_1, C_2) = \int_\Omega g\delta_\varepsilon(\phi)|\nabla\phi|\,dx + \mu\sum_{i=1}^{2}\int_\Omega \left(\int_\Omega K_\sigma(x-y)(C_i(x) - f(y)\log C_i(x))M_i(\phi(y))dy\right)dx$$
$$+ v \cdot \int_\Omega \frac{1}{2}(|\nabla\phi(x)| - 1)^2\,dx \tag{3}$$

Where $M_1 = H(\phi)$, $M_2 = 1 - H(\phi)$, $H(\phi)$ is approximated by a smooth function $H_\varepsilon(\phi)$ defined by

$$H_\varepsilon(\phi) = \frac{1}{2}\left[1 + \frac{2}{\pi}\arctan\left(\frac{\phi}{\varepsilon}\right)\right] \tag{4}$$

The derivative of $H_\varepsilon(\phi)$ is

$$\delta_\varepsilon(\phi) = H_\varepsilon'(\phi) = \frac{1}{\pi}\left(\frac{\varepsilon}{\varepsilon^2 + \phi^2}\right) \tag{5}$$

Where $g$ is a positive and decreasing edge detector function which is often defined as $g = \frac{1}{1 + \beta|\nabla f_\sigma \otimes I|^2}$, and $g$ usually takes smaller values at object boundaries than at other locations; The parameter $\beta$ controls the details of the segmentation, and $\nabla f_\sigma \otimes I$ is used to smooth the image to reduce the noise; In order to efficiently smooth multiplicative noise, we use the infinite symmetric exponential filter (ISEF) $f_\sigma(x) = \frac{1}{2\sigma}e^{-\frac{|x|}{\sigma}}$ with standard deviation $\sigma$ and of size $15 \times 15$ in edge detector $g$.

For fixed level set function $\phi$ , we minimize the function $E(\phi, C_1, C_2)$ with respect to the constant $C_1$ and $C_2$ . By calculus of variations, it is easy to solve them by

$$C_1 = \frac{K_\sigma(x) * (H_\varepsilon(\phi(x)) f(x))}{K_\sigma(x) * H_\varepsilon(\phi(x))}, C_2 = \frac{K_\sigma(x) * ((1 - H_\varepsilon(\phi(x))) f(x))}{K_\sigma(x) * (1 - H_\varepsilon(\phi(x)))} \tag{6}$$

Minimization of the energy functional $E(\phi, C_1, C_2)$ with respect to $\phi$ , can be obtained by solving gradient descent flow equation:

$$\frac{\partial \phi}{\partial t} = \delta_\varepsilon(\phi) div(g \frac{\nabla \phi}{|\nabla \phi|}) - \mu \cdot \delta_\varepsilon(\phi) \cdot \eta + v \cdot (\nabla^2 \phi - div(\frac{\nabla \phi}{|\nabla \phi|})) \tag{7}$$

Where $\eta = \int_\Omega K_\sigma(x-y)(C_1(y) - f(x)\log C_1(y))dy - \int_\Omega K_\sigma(x-y)(C_2(y) - f(x)\log C_2(y))dy$ .

**B. reaction-diffusion question [13]**

We have known chemical reactions or other processes governed by the reaction-diffusion equation for a vector $\phi(\mathrm{x,t},\xi)$ in a domain $\Omega$ in $R^n$

$$\phi_t = \xi \Delta \phi - \xi^{-1} V(\phi) \tag{8}$$

Here $\xi$ is a small parameter, which indicates that the reaction rate $\xi^{-1} V(\phi)$ is large while the diffusion coefficients is small.

We assume that $\phi$ also satisfies the initial and boundary conditions.

$$\phi(\mathrm{x,t}=0,\varepsilon) = \phi_0(x), \mathrm{x} \in \Omega; \partial_n \phi = 0, \mathrm{x} \in \partial\Omega \tag{9}$$

It has shown that[13], because the reaction rate is large and diffusion coefficients is small, the solution at each point $x$ will tend quickly to a stable equilibrium value of the reaction process, i.e., to a minimum of $V(\phi)$ . If $V(\phi)$ has a unique global minimum $\phi_t$ , then $\phi$ will tend to $\phi_t$ at every point $x$ , and the effect of diffusion will be to alter slightly the rate of approach.

**C. reaction-diffusion based variational Level Set method (RDLS)**

By adding a diffusion term into the level set evolution (LSE) equation of the proposed model (7), we construct a reaction-diffusion (RD) equation, which can gradually regularize the level set function (LSF) to be piecewise constant in each segment domain and gain the stable solution.

$$\phi_t = \xi\Delta\phi + \xi^{-1}[\delta_\varepsilon(\phi)div(g\frac{\nabla\varphi}{|\nabla\varphi|})\text{-}\mu\delta_\varepsilon(\phi)\eta] \qquad (10)$$

We can obtain the following energy functional $E(\phi,C_1,C_2,\xi)$ whose gradient flow is: $E(\phi,C_1,C_2,\xi)=\frac{1}{2}\xi\int_\Omega|\nabla\phi|^2dx+\frac{1}{\xi}\int_\Omega E(\phi,C_1,C_2)dx$ (11)

Equation (11) can be computed by (12) and (13):

$$\phi^{n+1/2}=\phi^n+(\Delta t_1\frac{1}{\zeta})\cdot\delta_\varepsilon(\phi)(div(g\frac{\nabla\phi}{|\nabla\phi|})\text{-}\mu\cdot\eta) \qquad (12)$$

$$\phi^{n+1}=\phi^n+(\Delta t_2\zeta)\cdot\Delta\phi^n \qquad (13)$$

Where $\phi^n=\phi^{n+1/2}$, $\Delta t_1\leftarrow\Delta t_1\cdot\frac{1}{\zeta}$, $\Delta t_2\leftarrow\Delta t_2\cdot\zeta$.

However, the energy functional (11) also do not have a unique global minimizer because they are all homogeneous degree one. By setting $\delta_\varepsilon(\phi)=1$ and $v=0$, we can get the stationary solutions of (11) as follows:

$$\frac{\partial\phi}{\partial t}=div(g\frac{\nabla\phi}{|\nabla\phi|})-\mu\cdot\eta \qquad (14)$$

The simplied flow represent gradient descent for minimizing the two energy functionals:

$$E(\phi)=\|\nabla\phi\|_g+\mu<\phi,\eta> \qquad (15)$$

We suppose the weighted TV term of the energy functionals (15) is anisotropic. By applying the anisotropic TV term to (15) and restricting the solution $\phi$ to lie in a finite interval ( $0\le\phi\le1$ ), we can obtain the anisotropic TV-based image globally convex segmentation model as follows:

$$\min_{0\le\phi\le1}\left\|\nabla_x\phi\right\|_g+\left\|\nabla_y\phi\right\|_g+\mu<\phi,\eta>+\frac{\alpha}{2}\left\|\phi-\frac{1}{2}\right\|^2 \tag{16}$$

We introduce the third term in (16) with $\alpha>0$ so that (16) is strictly convex and hence the solution is unique. The constant is introduced in the third term so that the minimizer does not bias towards 0 or 1.

## Ⅲ. FAST GLOBALLY CONVEX ALGORITHMS FOR SOLVING THE RDLS MODEL

### A. Split Bregman method for RDLS model

In the past, solutions of the TV model were based on nonlinear partial differential equations and the resulting algorithms were very complicated. It is well known that the split Bregman method is a technique for fast minimization of $l^1$ regularized energy functional. A break through was made by Goldstein-Osher (GO) [15-16].They proposed a more efficient way to compute the TV term based on the split Bregman method. To apply the split Bregman method to (16), we introduce auxiliary variables $d_x\leftarrow\nabla_x\phi$, $d_y\leftarrow\nabla_y\phi$, and add a quadratic penalty function to weakly enforce the resulting equality constraint which results in the following unconstrained problem:

$$\begin{aligned}&(\phi^{k+1},d_x^{k+1},d_y^{k+1})=\arg\min_{0\le\phi\le1}\left\|d_x\right\|_g+\left\|d_y\right\|_g+\mu<\phi,\eta^k>+\\&\frac{\lambda}{2}\left\|d_x-\nabla_x\phi\right\|^2+\frac{\lambda}{2}\left\|d_y-\nabla_y\phi\right\|^2+\frac{\alpha}{2}\left\|\phi-\frac{1}{2}\right\|^2\end{aligned} \tag{17}$$

We then apply split Bregman method to strictly enforce the constraints $d_x=\nabla_x\phi$, $d_y=\nabla_y\phi$. The resulting optimization problem becomes:

$$\begin{aligned}&(\phi^{k+1},d_x^{k+1},d_y^{k+1})=\arg\min_{0\le\phi\le1}\left\|d_x\right\|_g+\left\|d_y\right\|_g+\mu<\phi,\eta^k>+\\&\frac{\lambda}{2}\left\|d_x-\nabla_x\phi-b_x^k\right\|_2^2+\frac{\lambda}{2}\left\|d_y-\nabla_y\phi-b_y^k\right\|_2^2+\frac{\alpha}{2}\left\|\phi-\frac{1}{2}\right\|^2\\&b_x^{k+1}=b_x^k+\nabla_x\phi^{k+1}-d_x^{k+1}\\&b_y^{k+1}=b_x^k+\nabla_y\phi^{k+1}-d_y^{k+1}\end{aligned} \tag{18}$$

For fixed $\vec{d}$, the Euler-Lagrange equation of optimization problem (18) with respect to $\phi$ is:

$$(\alpha I-\lambda\Delta)\phi^{k+1}=\frac{\alpha}{2}-\mu\cdot\eta^{k}+\lambda\cdot\nabla_x^T(d_x^k-b_x^k)+\lambda\cdot\nabla_y^T(d_y^k-b_y^k) \tag{19}$$

For fixed $\phi$, minimization of (18) with respect to $\vec{d}$ gives:

$$\begin{aligned} d_x^{k+1}&=shrink_{g/\lambda}(\nabla_x\phi^{k+1}+b_x^k)\\ d_y^{k+1}&=shrink_{g/\lambda}(\nabla_x\phi^{k+1}+b_y^k)\end{aligned} \tag{20}$$

Where $shink_{g/\lambda}(x)=\operatorname{sgn}(x)\max(|x|-g/\lambda,0)$.

By using central discretization for Laplace operator and backward difference for divergence operator, the numerical scheme for (19) becomes:

$$\begin{aligned} &\alpha_{i,j}=d_{i\text{-}1,j}^{x}-d_{i,j}^{x}-b_{i-1,j}^{x}+b_{i,j}^{x}+d_{i,j-1}^{y}-d_{i,j}^{y}-b_{i,j-1}^{y}+b_{i,j}^{y}\\ &\beta_{i,j}=\frac{\lambda}{\alpha+4\lambda}(\phi_{i-1,j}+\phi_{i+1,j}+\phi_{i,j-1}+\phi_{i,j+1}+\alpha_{i,j})+\frac{0.5\alpha-\mu\cdot\eta^{k}}{\alpha+4\lambda}\\ &\phi_{i,j}=\max(\min(\beta_{i,j},1),0)\end{aligned} \tag{21}$$

As the optimal $\phi$ is found, the segmented region can be found by thresholding the level set function $\phi(x)$ for some $\gamma\in(0,1)$ : $\Omega_1=\{x:\phi(x)>\gamma\}$ . The split Bregman algorithm for the minimization problem (18) can be summarized as follows:

---

**Split Bregman method for anisotropic TV –based RDLS Model (SBRD)**

Given: noisy image $f$; $\lambda>0$, $\mu>0$

Initialization: $b^0=0$, $d^0=0$, $\phi^0=f/\max(f)$

For $k=0,1,2,\cdots$

Comute (21)

$\phi^{k+1}=\max(\min(\phi^{k},1),0)$

$d_x^{k+1}=shrink_{g/\lambda}(\nabla_x\phi^{k+1}+b_x^k)$

$d_y^{k+1}=shrink_{g/\lambda}(\nabla_x\phi^{k+1}+b_y^k)$

$b_x^{k+1}=b_x^k+\nabla_x\phi^{k+1}-d_x^{k+1}$

$b_y^{k+1}=b_x^k+\nabla_y\phi^{k+1}-d_y^{k+1}$

END

---

**B. Fixed point algorithm 1 for for RDLS model(FPRD1)**

We know the energy functional of image segmenatation (16) does not have a unique global minimizer because it is homogeneous degree one. However, the ROF denoising model always admits a unique solution because the energy functional is strictly convex. In order to utilizing the convexity of ROF denoising model and favor segmentation along curves where the edge detector function is minima, we give a discrete version weighted ROF (WROF) denoising model based on anisotropic TV term as follows:

$$E_{WROF}(\phi) = \|\nabla_x \phi\|_g + \|\nabla_y \phi\|_g + \frac{\alpha}{2}\|\phi - f\|^2 \tag{22}$$

Where $\alpha > 0$ is an appropriately chosen positive parameter.

Note that GO algorithm [15-16] still requires solving a partial difference equation in each iteration step. Recently, Jia-Zhao (JZ)[17-18] proposed a fast algorithm for image denoising based on TV term, which is very simple and does not involve partial differential equations or difference equation. In order to applying the JZ algorithm to image segmentation question (16), we first suppose that $\phi^k$ and $\eta^k$ are known and reformulate (16) by adding a proximity term $\frac{\alpha}{2}\|\phi - \phi^k\|^2$ as[27,29,30]:

$$\begin{aligned}
\phi^{k+1} &= \arg\min_{0\le\phi\le1} \|\nabla_x \phi\|_g + \|\nabla_y \phi\|_g + \mu < \phi, \eta^k > + \frac{\alpha}{2}\|\phi - \phi^k\|^2 \\
&= \arg\min_{0\le\phi\le1} \|\nabla_x \phi\|_g + \|\nabla_y \phi\|_g + \mu < \phi - \phi^k, \eta^k > + \frac{\alpha}{2}\|\phi - \phi^k\|^2 \\
&= \arg\min_{0\le\phi\le1} \|\nabla_x \phi\|_g + \|\nabla_y \phi\|_g + \frac{\alpha}{2}\left\|\phi - (\phi^k - \frac{\mu\eta^k}{\alpha})\right\|^2
\end{aligned} \tag{23}$$

Treating $\phi^k - \frac{\mu\eta^k}{\alpha}$ as the **"noisy image"**, (23) minimizes the residue between the **"clean image"** $\phi$ and $\phi^k - \frac{\mu\eta^k}{\alpha}$ using the total variation norm, then (23) is the total variation denoising problem. So we propose a fixed point algorithm 1(FPA1) based on JZ algorithm to solve (23) as follows:

$$
\begin{aligned}
&b_x^k = (I - shrink_{g/\lambda})(\nabla_x \phi^k + b_x^{k-1}) \\
&b_y^k = (I - shrink_{g/\lambda})(\nabla_y \phi^k + b_y^{k-1}) \\
&\phi^{k+1} = \phi^k - \frac{\mu\eta^k}{\alpha} - \frac{\lambda}{\alpha}(\nabla_x^T b_x^k + \nabla_y^T b_y^k) \\
&\phi^{k+1} = \max(\min(\phi^{k+1}, 1), 0)
\end{aligned} \tag{24}
$$

According to [23-24], we infer from (24) that operator $(I - shrink_{g/\lambda})(I - \frac{\lambda}{\alpha}\nabla\nabla^T)$ is nonexpansive when $\frac{\lambda}{\alpha}$ is less than $\frac{1}{4}\sin^{-2}\frac{(N-1)\pi}{2N}$ which is slightly bigger than 1/4 .In order to accelerate the convergence of (23) , we adopt the following iteration scheme by utilizing k-averaged operator theory ( see more details in [24]):

$$
\begin{aligned}
&b_x^k = t \cdot b_x^{k-1} + (1-t)\cdot(I - shrink_{g/\lambda})(\nabla_x \phi^k + b_x^{k-1}) \\
&b_y^k = t \cdot b_y^{k-1} + (1-t)\cdot(I - shrink_{g/\lambda})(\nabla_y \phi^k + b_y^{k-1})
\end{aligned} \tag{25}
$$

Where the weight factor $t \in (0,1)$ is called the relaxation parameter. As the optimal $\phi(x)$ is found, the segmented region can be found by thresholding the function $\phi(x)$ for some $0 < \gamma < 1$ : $\Omega_1 = \{x : \phi(x) > \gamma\}$ . The FPA1 for the minimization problem (23) can be summarized as follows:

---

FPA1 for anisotropic TV–based RDLS Model:

Given: noisy image $f$ ; $\lambda > 0$ , $\mu > 0$ , $\alpha > 0$ , $t \in (0,1)$ , $\gamma \in (0,1)$

Initialization: $b_x^0 = 0$ , $b_y^0 = 0$ , $\phi^0 = f / \max(f)$ , $\Omega_0 = \{x : \phi^0 > \gamma\}$ ,he

$C_1^0 = \int_{\Omega_0} f dx$ , $C_2^0 = \int_{\Omega_0^c} f dx$

For $k = 0,1,2,\cdots$

$b_x^k = t \cdot b_x^{k-1} + (1-t)\cdot(I - shrink_{g/\lambda})(\nabla_x \phi^k + b_x^{k-1})$

$b_y^k = t \cdot b_y^{k-1} + (1-t)\cdot(I - shrink_{g/\lambda})(\nabla_y \phi^k + b_y^{k-1})$

$\phi^{k+1} = \phi^k - \frac{\mu\eta^k}{\alpha} - \frac{\lambda}{\alpha}(\nabla_x^T b_x^k + \nabla_y^T b_y^k)$

$\phi^{k+1} = \max(\min(\phi^{k+1}, 1), 0)$

$\Omega_{k+1} = \{x : \phi(x) > \gamma\}$ , $C_1^{k+1} = \int_{\Omega_{k+1}} f dx$ , $C_2^{k+1} = \int_{\Omega_{k+1}^c} f dx$

END

---

**C. Fixed point algorithm 2 for for RDLS model(FPRD2)**

We can also reformulate (16) by introducing a term $\frac{\alpha}{2}\|\phi - \varphi\|^2$ as:

$$(\phi^{k+1},\varphi^{k+1}) = \arg\min_{\phi,0\le\varphi\le1} \|\nabla_x \phi\|_g + \|\nabla_y \phi\|_g + \mu < \varphi,\eta > + \frac{\alpha}{2}\|\phi-\varphi\|^2 \tag{26}$$

We then apply split Bregman method to (26) and propose a fixed point algorithm 2 (FPA2) as follows:

$$\begin{aligned}
&\phi^{k+1} = \arg\min_{0\le\phi\le1} \|\nabla_x \phi\|_g + \|\nabla_y \phi\|_g + \frac{\alpha}{2}\|\phi-\varphi^k - c^k\|^2 \\
&\varphi^{k+1} = \arg\min_{0\le\varphi\le1} \frac{\alpha}{2}\|\phi^{k+1}-\varphi - c^k\|^2 + \mu < \varphi,\eta^k > \\
&c^{k+1} = c^k + \varphi^{k+1} - \phi^{k+1}
\end{aligned} \tag{27}$$

we can get a solution of (26) as FPA1,i.e.

$$\begin{aligned}
&\phi^{k+1} = \varphi^k + c^k - \frac{\lambda}{\alpha}(\nabla_x^T b_x^k + \nabla_y^T b_y^k) \\
&\varphi^{k+1} = \phi^{k+1} - c^k - \frac{\mu}{\alpha}\eta^k, \\
&c^{k+1} = c^k + \varphi^{k+1} - \phi^{k+1} \\
&\varphi^{k+1} = \max(\min(\varphi^{k+1},1),0)
\end{aligned} \tag{28}$$

Where $b_x^k$ and $b_y^k$ is also defined as (25). As the optimal $\varphi$ is found, the segmented region can be found by thresholding the function $\varphi(x)$ for some $\gamma \in (0,1)$ : $\Omega_1 = \{x:\varphi(x) > \gamma\}$ . The FPA2 for the minimization problem (27) can be summarized as follows:

---

FPA2 for anisotropic TV –based RDLS Model:

Given: noisy image $f$ ; $\lambda > 0$ , $\mu > 0$ , $\alpha > 0$ , $t \in (0,1)$ , $\gamma \in (0,1)$

Initialization: $b_x^0 = 0$ , $b_y^0 = 0$ , $c^0 = 0$ , $\phi^0 = f/\max(f)$

$\Omega_0 = \{x:\phi^0 > \gamma\}$ , $C_1^0 = \int_{\Omega_0} f dx$ , $C_2^0 = \int_{\Omega_0^c} f dx$

For $k = 0,1,2,\cdots$

$b_x^k = t \cdot b_x^{k-1} + (1-t)\cdot(I - shrink_{g/\lambda})(\nabla_x \phi^k + b_x^{k-1})$

$b_y^k = t \cdot b_y^{k-1} + (1-t)\cdot(I - shrink_{g/\lambda})(\nabla_y \phi^k + b_y^{k-1})$

$\phi^{k+1} = \varphi^k + c^k - \frac{\lambda}{\alpha}(\nabla_x^T b_x^k + \nabla_y^T b_y^k)$

$\varphi^{k+1} = \phi^{k+1} - c^k - \frac{\mu}{\alpha}\eta^k,$

$c^{k+1} = c^k + \varphi^{k+1} - \phi^{k+1}$

$\varphi^{k+1} = \max(\min(\varphi^{k+1},1),0)$

$\Omega_{k+1} = \{x:\varphi(x) > \gamma\}$ , $C_1^{k+1} = \int_{\Omega_{k+1}} f dx$ , $C_2^{k+1} = \int_{\Omega_{k+1}^c} f dx$

END

---

**D. Segmentation accuracy**

We adopt the uniformity measurement of image segmentation regions to evaluate the performance of the proposed method. The interior of each region should be uniform after the segmentation and there should be a great difference among different regions. That is to say, the uniformity degree of regions represents the quality of the segmentation. Therefore, we give the measurement of segmentation accuracy (SA) as follows [26]:

$$pp = 1 - \frac{1}{C}\sum_i \{\sum_{x\in R_i}[f(x) - \frac{1}{A_i}\sum_{x\in R_i} f(x)]^2\} \quad (29)$$

Where $R_i$ denotes different segmentation regions, $C$ is the normalization constant, $f(x)$ is the gray value of point $x$ in the image, $A_i$ is the number of the pixels in each region $R_i$. The closer to 1 the value of $pp$ is, the more uniform the interior of the segmentation regions are and the better the quality of the segmentation is.

## Ⅳ. EXPERIMENTAL RESULTS

All the algorithms are implemented with Matlab 8.0 in core2 with 1.9 GHZ and 1GB RAM. We download SAR image1 and image2 from the web: http://www.eorc.jaxa.jp/ JERS-1/en/GFMP/SEA-2A, which have the size $306\times332$ and the gray-scale in the range between 0 and 255, and get SAR image3 from http://www.sandia.gov/radar/datacoll.html, which has the size $256\times141$ and the gray-scale in the range between 0 and 255.

We adopt the following parameters. The parameters of RDLS model are chosen as $\mu = 15$ , $\beta = 20$ $\varepsilon = 1$ , $\Delta t_1 = 0.1$ , $\Delta t_2 = 0.15$ , $\sigma = 15$ .The parameters of SBRD algorithm are chosen as $\lambda = 1000$ , $\mu = 0.006\lambda$ , $\beta = 20$ , $\varepsilon = 1$ , $\sigma = 15$ (also can use a little one such as 8).The parameters of FPRD1 is chosen as $\mu = 0.15$ , $\lambda = 1$ , $\alpha = 12, \beta = 20, \varepsilon = 1, \sigma = 15$ , $t = 1e-4$ . The parameters of FPRD2 is $\mu = 0.1, \lambda = 1$ , $\alpha = 8, \beta = 12, \varepsilon = 1, \sigma = 15$ , $t = 1e-4$ . The thresholding values for SBA and FPRD1 and FPRD2 are all chosen as $\gamma = 0.5$ , which are used to find the segmented region $\Omega_1 = \{x : u(x) > \gamma\}$ . We show the segmentation results of three SAR images in Fig.1-Fig.3.

We compare image segmentation speed (the pair (.,.)) and $pp$ values between RDLS,SBRD,FPRD1 and FPRD2 in TABLE Ⅰ. We observe that the $pp$ values of SBRD ,FPRD1 and FPRD2 are all close to 1, which show high precision of the proposed fast algorithm. SBA , FPRD1 and FPRD2 all give good results and are robustness to initialization contour.The proposed SBRD model can reduce greatly the time needed for the RDLS model. Iterative time required for the proposed FPRD1/FPRD2 algorithms are about 1/2 (or less than) of the time required for the SBRD algorithm based on the Split Bregman technique.

## Ⅴ.CONCLUSION

In this paper, we propose a novel locally statistical variational active contour model based on I-divergence-TV denoising model. By adding a diffusion term into the level set evolution (LSE) equation of the proposed model, we construct a reaction-diffusion (RD) equation, which can gradually regularize the level set function (LSF) to be piecewise constant in each segment domain and gain the stable solution. We transform the proposed model into classic ROF model by adding a proximity term. Inspired by a fast denoising algorithm proposed by Jia-Zhao , we propose two fast fixed point algorithms to solve SAR image segmentation question. Experimental results for real SAR images show that the proposed image segmentation model can efficiently stop the contours at weak or blurred edges, and can automatically detect the exterior and interior boundaries of images with multiplicative gamma noise. The proposed FPRD1/FPRD2 models are about 1/2 (or less than) of the time required for the SBRD model based on the Split Bregman technique.

The reason why the proposed algorithms are faster than the SBRD algorithm are that they do not involve partial differential or difference equations. The proposed algorithms can also be applied to the case of isotropic TV, and all the segmenation question are solved by the ROF model which has been intensively studied.

## REFERENCES

[1] M. Silveira, and S. Heleno, "Separation between water and land in SAR images using region-based level sets," IEEE Geoscience and Remote Sensing Letters, Vol.6, no.3, pp.471-475, July, 2009.

[2] M.Horritt, "A statistical active contour model for SAR image segmentation," Image Vis. Comput., vol.17, no.3, pp.213-224, Mar.,1999.

[3] B.Huang, H.Li, and X.Huang, "A level set method for oil slick segmentation in SAR images," Int. J. Remote Sens., vol.26, no.6, pp.1145-1156, Mar., 2005.

[4] I.Ben Ayed, A.Mitiche, and Z.Belhadj, "Multiregion level-set partitioing of synthetic aperture radar images," IEEE Trans. Pattern Anal. Mach. Intell., vol.27, no.5, pp. 793-800, May, 2005.

[5] T.Chan and L.Vese, "Active contours without edges," IEEE Trans Image Process., vol.10, no.2, pp.266-277, Feb. ,2001.

[6] L.Vese and T.Chan, "A multiphase level set framework for image segmentation using the Mumford and Shah model," Int. J. Comput.Vis., vol.50, pp.271-293, 2002.

[7] C. Li, C.Kao, J.Gore, Z.Ding,, "Minimization of region-scalable fitting energy for image segmentation," IEEE Trans.Imag.Proc, vol.17,no.10, pp.1940-1949, Oct.,2008.

[8] N.Paragios and R.Deriche, "Geodesic active regions and level set methods for supervised texture segmentation," Int. J. Comput. Vis., vol.46, no.3, pp.223-247, 2002.

[9] V.Caselles, R.Kimmel, and G.Sapiro, "Geodesic active contours," Int. J. Comput. Vis., vol.22, no.1, pp.61-79, Feb.,1997.

[10] M.Kass, A.Witkin, and D.Terzopoulos, "Snakes: active contour models," Int. J. Comput. Vis., vol.1, no.4, pp.321-331, 1988.

[11] S.Kichenassamy, A.Kumar, P.Olver, A.Tannenbaum, and A.Yezzi, "Gradient flows and geometric active contour models," In Proc. 5th Int Conf. Computer Vision, pp.810-815, 1995.

[12] C.Li, C.Xu, C.Gui, and M.D.Fox, "Level set evolution without re-initialization: A new variational formulation," In Proc. IEEE Conf. Computer Vision and Pattern Recognition, vol.1, pp.430-436, 2005.

[13] Kaihua Zhang,Lei Zhang etal, "Re-initialization Free Level Set Evolution via Reaction Diffusion," IEEE Trans.Imag.Proc, vol.22,no.1, pp.258-271, January,2013.

[14] T.Chan, S.Esedoglu, and M.Nikolova, "Algorithms for finding global minimizers

of image segmentation and denoising models," SIAM J. Appl. Math., vol.66, no.5, pp.1632-1648, 2006.

[15]T.Goldstein and S.Osher, "The split Bregman algorithm for L1 regularized problems," SIAM J. Imaging Sci., vol.2, no.2, pp.323-34, 2009.

[16] T.Goldstein, X.Bresson, and S.Osher, "Geometric applications of the split Bregman method:segmentation and surface reconstruction," J. Sci. Comput., vol.45, no.1-3,pp.272-293, 2010.

[17]R.-Q.Jia, and H.Zhao, "A fast algorithm for the total variation model of image denoising," Adv. Comput. Math., vol.33, no.2, pp.231-241, 2010.

[18]R.-Q.Jia, H.Zhao and W.Zhao, "Relaxation methods for image denoising based on difference schemes," SIAM J. Appl. Math., vol.9, no.1, pp.355-372, 2011.

[19] L.Rudin, S.Osher, and E.Fatemi, "Nonlinear total variation based noise removal algorithms," Phys. D, vol.60, no.1-4, pp.259-268, 1992.

[20] G.Steidl and T.Teuber, "Removing multiplicative noise by Douglas-Rachford splitting methods," J. Math. Imaging Vi., vol.36,no. 2, pp.168-184, 2010.

[21]K.-H.Zhang, L.Zhang, H.Song, W.-G.Zhou, "Active contours with selective local or global segmentation: A new formulation and level set method," Image and Vision computing, vol.28,no.4,pp.668-676,2010.

[22] V.Caselles, A.Chambolle, and M.Novaga, "The discontinuity set of solutions of the TV denoising problem and some extensions," Multiscale Model. Simul., vol6, pp.879-894, 2007.

[23]C.A.Micchelli, L.Shen and Y.Xu, "Proximity algorithms for image models: denoising," Inverse Probl. vol.27,no.4, 2011.

[24]P.Combettes and V.Wajs, "Signal recovery by proximal forward-backward splitting," Multiscale Model. Simul. vol4, no.4, pp.1168-1200, 2006.

[25]X.Bresson, S.Esedoglu, P.Vandergheynst, J.-P.Thiran and S.Osher, “Fast global minimization of the active contour/snake model,” J. Math. Imaging Vis. Vol.28, pp.151-167,2007.

[26]T.D.Ross and J.C.Mossing, “The MSTAR evalution methodology,” Proceeding of SPIE, 3721, pp.705-713, 1999.

[27] LIU Guangming, MENG Xiangwei. A novel SAR image locally statistical active contour model and algorithm[J]. Geomatics and Information Science of Wuhan University, vol.40,no.5,pp.628-631,2015.

[28]Guangming Liu et al. A novel SAR image local fitting active contour model. Fire Control Radar Technology,vol.43,no.1,pp.5-8,2014.

[29]Guangming Liu etal. A global optimization SAR image segmentation model can be easily transformed to a general ROF denoising model. arXiv:2312.08376[cs.CV],Dec.202.

[30]Guangming Liu etal. SAR image segmentation algorithms based on I-divergence-TV model. arXiv:2312.09365[cs.CV],Dec.2023.

[31] S. V. Venkatakrishnan, C. A. Bouman, and B. Wohlberg, “Plugand-play priors for model based reconstruction,” in IEEE Global Conference on Signal and Information Processing, 2013, pp. 945–948.

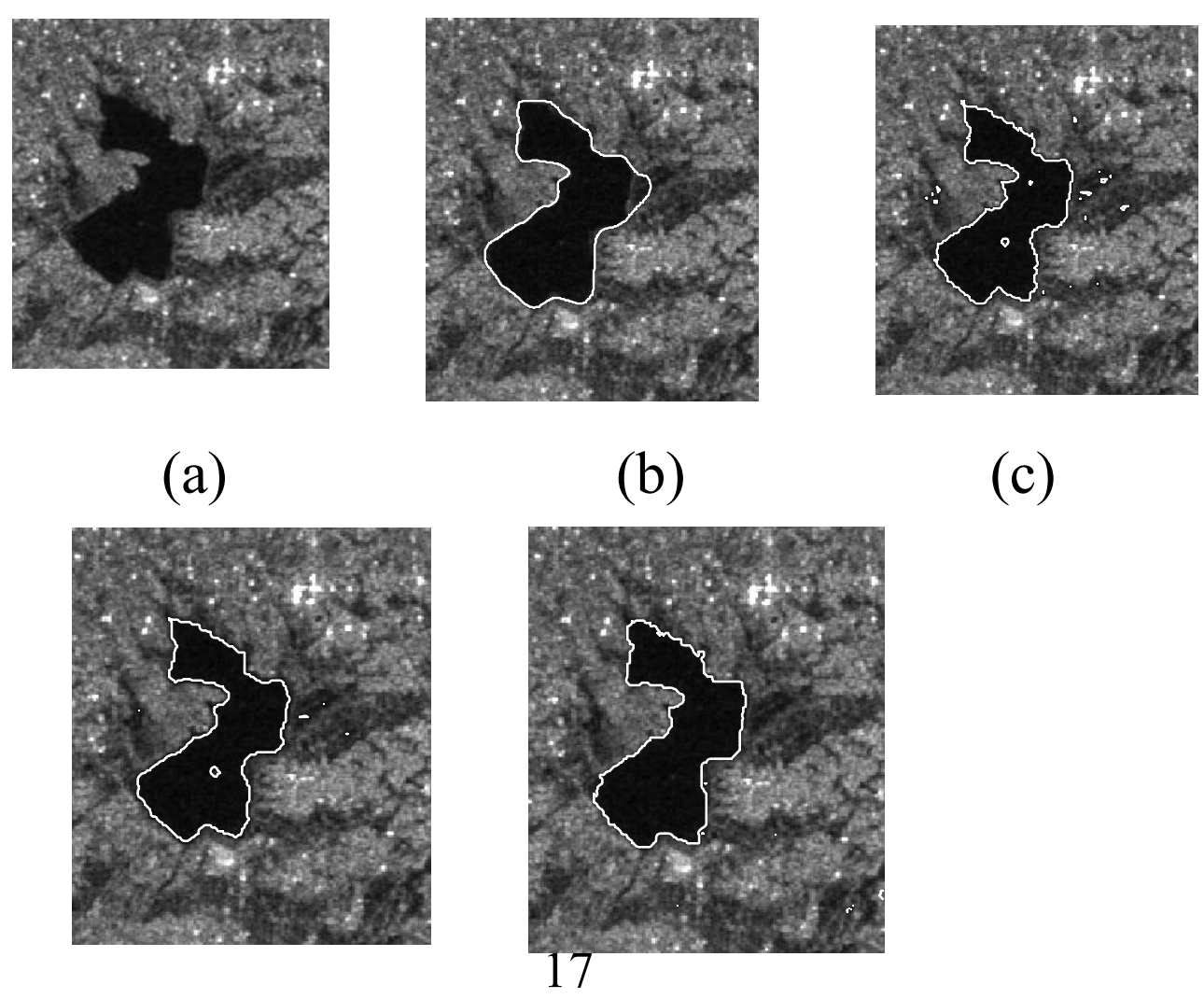

(a) (b) (c)

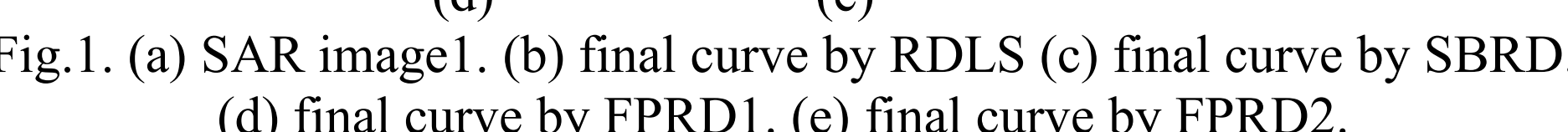

(d) (e)

Fig.1. (a) SAR image1. (b) final curve by RDLS (c) final curve by SBRD. (d) final curve by FPRD1. (e) final curve by FPRD2.

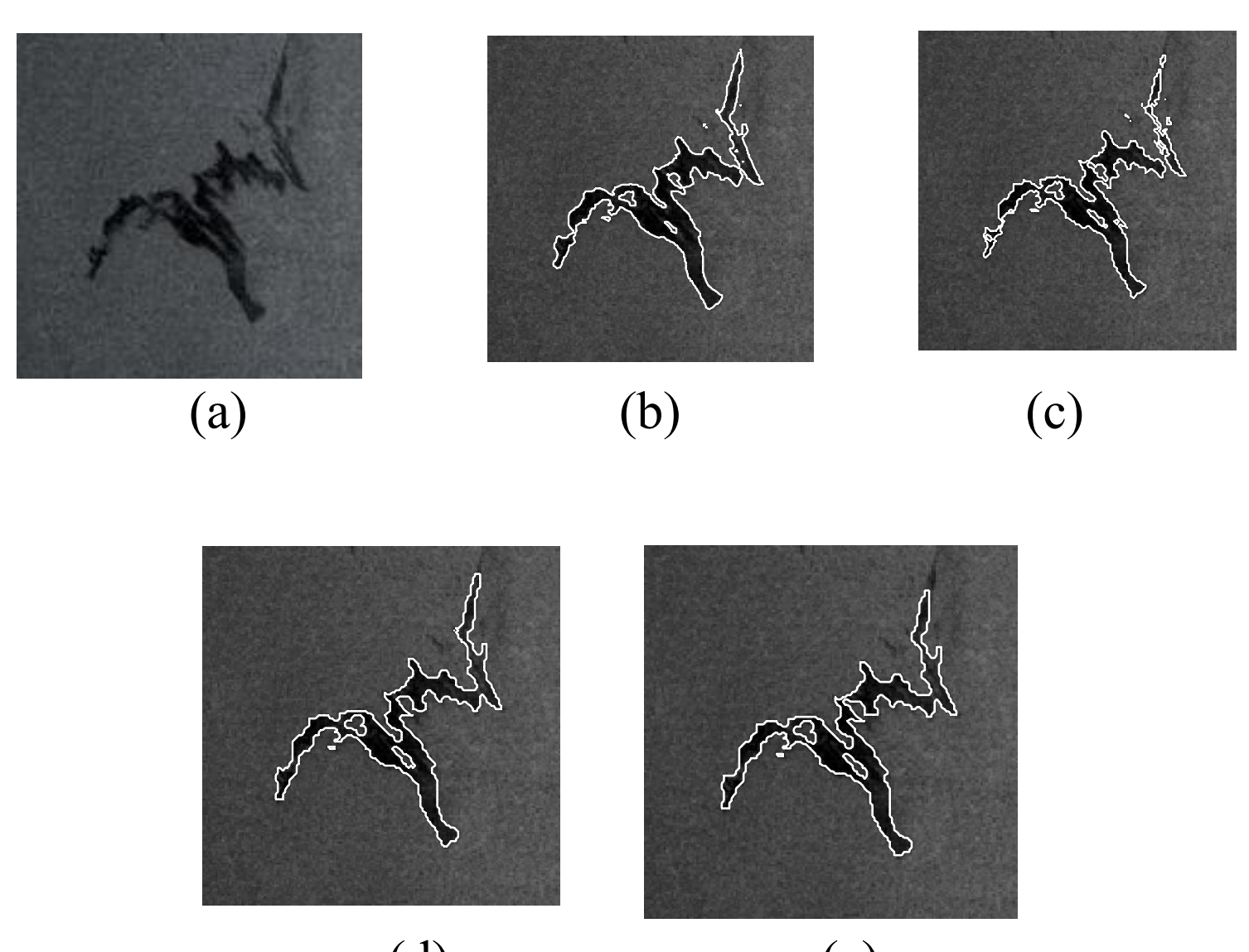

(a) (b) (c)

(d) (e)

Fig.2. (a) SAR image2. (b) final curve by RDLS (c) final curve by SBRD. (d) final curve by FPRD1. (e) final curve by FPRD2.

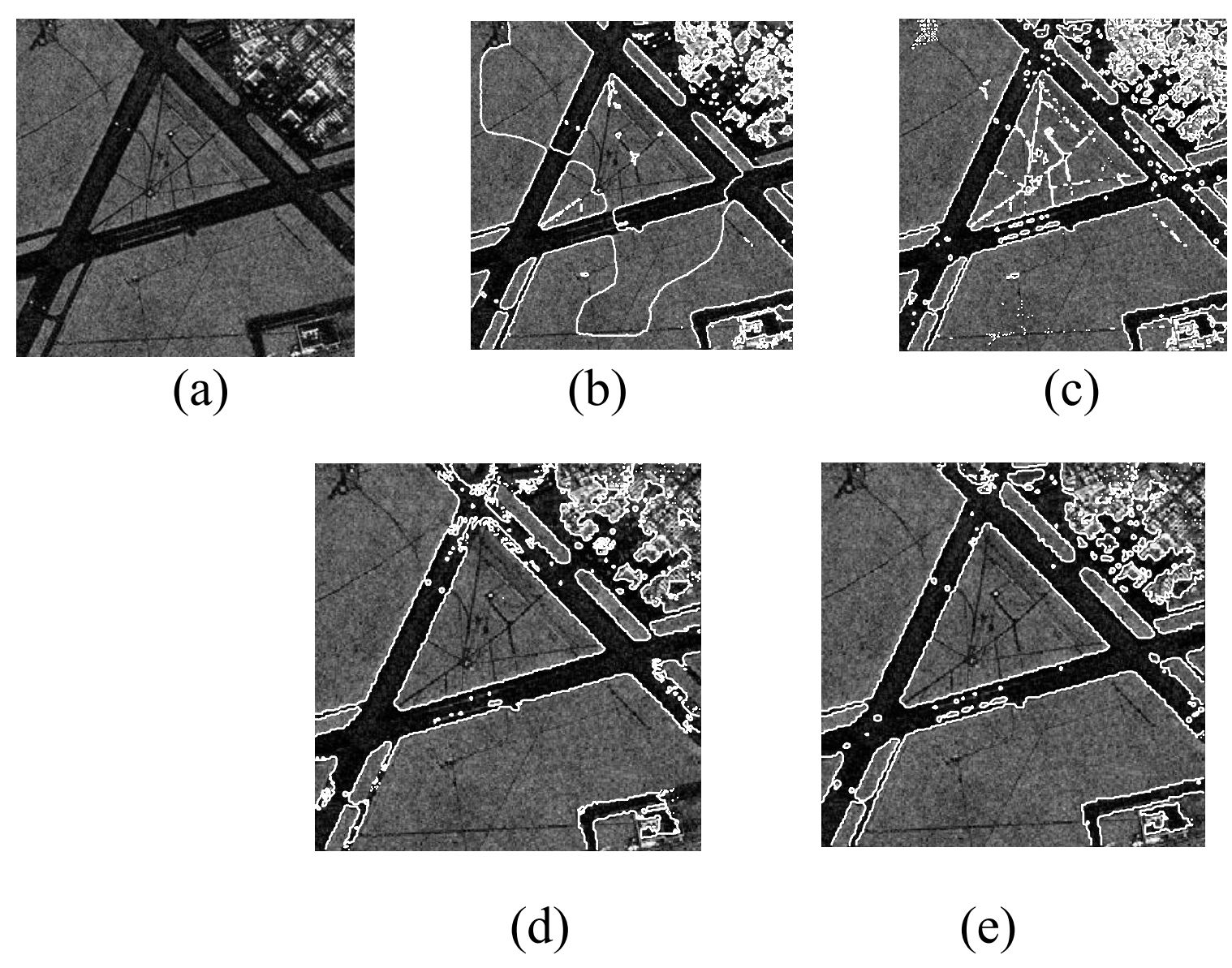

(a) (b) (c)

(d) (e)

Fig.3. (a) SAR image3. (b) final curve by RDLS (c) final curve by SBRD.(d) final curve by FPRD1. (e) final curve by FPRD2.

TABLE Ⅰ

SPEED AND SA OF RDLS,SBRD,FPRD1 AND FPRD2

| Image | RDLS | SBRD | FPRD1 | FPRD2 |
|---|---|---|---|---|
| 1 | (140,251.578s),0.9672 | (16,25.563s),0.993 | (60,13.142s),0.995 | (30,7.51s),0.991 |
| 2 | (130,118.891s),0.9654 | (25,20.219s),0.993 | (28,9.307s),0.994 | (20,8.65s),0.99 |
| 3 | (160,556.047s),0.9647 | (26,88.187s),0.995 | (34,17.956),0.996 | (32,16.434s),0.993 |